\documentclass{INTERSPEECH2023}

\title{PSST! Prosodic Speech Segmentation with Transformers}
\name{Nathan Roll$^1$, Calbert Graham$^2$, Simon Todd$^1$}
\address{
  $^1$University of California, Santa Barbara, USA\\
  $^2$University of Cambridge, UK}
\email{nroll@ucsb.edu, crg29.cam.ac.edu, sjtodd@ucsb.edu}

\begin{document}

\title{\textbf{PSST! Prosodic Speech Segmentation with Transformers}}

\author{\IEEEauthorblockN{Nathan Roll}
\IEEEauthorblockA{\textit{University of California, Santa Barbara} \\
\texttt{nroll@ucsb.edu}}
\and
\IEEEauthorblockN{Calbert Graham}
\IEEEauthorblockA{\textit{University of Cambridge} \\
\texttt{crg29@cam.ac.uk}}
\and
\IEEEauthorblockN{Simon Todd}
\IEEEauthorblockA{\textit{University of California, Santa Barbara} \\
\texttt{sjtodd@ucsb.edu}}
}

\maketitle

\begin{abstract}
\normalfont{Self-attention mechanisms have enabled transformers to achieve superhuman-level performance on many speech-to-text (STT) tasks, yet the challenge of automatic prosodic segmentation has remained unsolved. In this paper we finetune Whisper, a pretrained STT model, to annotate intonation unit (IU) boundaries by repurposing low-frequency tokens. Our approach achieves an accuracy of 95.8\%, outperforming previous methods without the need for large-scale labeled data or enterprise-grade compute resources. We also diminish input signals by applying a series of filters, finding that low pass filters at a 3.2 kHz level improve segmentation performance in out of sample and out of distribution contexts. We release our model\footnote{\url{https://github.com/Nathan-Roll1/PSST}} as both a transcription tool and a baseline for further improvements in prosodic segmentation.}

\end{abstract}


\section{Introduction}
Listeners are able to perceive speech as a coherent sequence of distinct sounds, despite the fact that speech is spoken in a continuous stream with coarticulatory effects in which the realization of phonemes is influenced by adjacent phonemes \cite{roach_phonetic_1990, white_segmentation_2018}. Automatic speech recognition (ASR) is the process by which a machine transforms audio into a sequence of words. This ability of humans to recognize patterns in speech is fundamental to the communication process. ASR entails automatically mapping linguistic categories such as words, syllables, or phones to the corresponding acoustic signal. ASR has various important applications (e.g., in virtual assistants and chatbots, voice commands and dictation, live captioning and transcriptions). Yet, despite significant technological progress, speech segmentation by machines has remained one of the most difficult challenges in speech processing. The main reason for this difficulty is that there are multiple sources of variation in speech, including speaker characteristics (e.g., age, gender, vocal tract length), the environment (e.g., microphone, room acoustics), speaking style (e.g., spontaneous vs planned), and so on. 

The linguistic analysis of speech has traditionally focused on the analysis of segments. However, a growing body of research focuses on prosody: the autosegmental structure of the utterance that encodes information about prominence and phrasal organization \cite{pierrehumbert_prosody_1999, ladd_intonational_2008}. Examples of prosody in speech include  interconnected and interacting phenomena  such as intonation, stress, rhythm, and phrasing \cite{arvaniti_phonetics_2020}. In English, phrasal organization serves to group words into chunks that are used by the listeners and speakers to process the utterance. A boundary partitions each of these chunks, which are important in enhancing speech intelligibility \cite{cooper_fundamental_1981,selkirk_phonology_1984}. This helps the listeners to correctly discern the syntactic structure of the utterance and deduce its meaning \cite{streeter_acoustic_1978,wingfield_prosodic_1984,beach_interpretations_1991,crystal_prosodic_1986,warren_prosody_1996}. The current study aims to develop an automatic intonation unit segmentation and boundary detection tool for American English.

Prosodic cues can play a significant role in spoken discourse processing of American English \cite{venditti_intonation_2003,hirschberg_communication_2002}. The naturalness of text-to-speech (TTS) systems relies, in part, on being able to generate pauses throughout the utterance in locations in which a speaker would naturally produce them. Some researchers have used durational cues and pauses to detect prosodic phrase boundaries \cite{yang_duration_2003,salomon_detection_2004}, while others have used multiple prosodic cues in automatic boundary detection \cite{mandal_word_2003,peters_multiple_2003}. However, we have not yet been able to fully explain prosody within ASR techniques, for a number of reasons: (1) significant speaker and contextual variation in prosodic realization, (2) the highly complex relations between prosodic structure and other levels of organization in an utterance, (3) the difficulty in separating pauses that are meant to indicate boundaries from those that result from unintentional errors. Humans overcome these challenges in the identification of IUs, also known as intonational phrases or prosodic units, through either express or implicit recognition of various, often subtle, cues. These include gestalt unity of intonation contour, pitch reset and anacrusis at unit onset, lag at unit terminus, pauses and breaths at unit boundaries, and nuclear accents \cite{du_bois_discourse_1992}. 

Automatic approaches to IU recognition have varied widely, from simple rule-based algorithms \cite{biron_automatic_2021} to complex supervised machine learning models \cite{stehwien_prosodic_2017}. The novelty of our method is as much theoretical as it is architectural. We view prosody not as a standalone problem, but as one strongly coupled with syntax. On the computational side, the success of transformer models in vanilla STT tasks, which represent half of the syntax-prosody interface \cite{bennett_syntaxprosody_2019}, yield a natural starting point for an end-to-end prosodic transcription application. 

In this paper, we investigate whether explicit supervision from a small, high quality dataset can “teach” a pretrained transformer-based STT model to segment speech into IUs. We focus on the following primary research objectives:

\begin{enumerate}
  \item To repurpose ASR-optimized transformer models to perform reliable IU boundary detection.
  \item To discover the role of sound versus syntax in such models via replication over diminished versions of the finetuning set.
  \item To test the robustness of prosodic boundary predictions through harmonic frequency filtering and evaluation of out-of-distribution speech data.
\end{enumerate}

\smallskip
\section{Methods}
\subsection{Data}\label{AA}
For training we use the Santa Barbara Corpus of Spoken American English (SBCSAE) for its breadth of participants and quality of transcription. The corpus contains spontaneous discourse and prosodically-annotated transcriptions from 60 conversations (210 individuals), spanning a total of $\sim$20 hours. The speakers vary in age from 11 to 101 years old and self-identify as Asian-American, Black/African-American, Latinx/Chicanx, Hispanic, Japanese, Native American, White, biracial, or other. They represent 30 U.S. states and educational backgrounds ranging from grade school to various post-graduate degrees. The corpus is roughly gender balanced, with 55\% of speakers identifying as female and 44\% as male. No gender data is available for the remaining speaker. SBCSAE transcriptions were performed by multiple trained examiners, with inconsistencies resolved by experts. All personal identifiers and otherwise sensitive pieces of information (as determined by the corpus creators) were masked using a 400 Hz low-pass filter with gradual fading in the 45 milliseconds before and after the region in question \cite{du_bois_john_w_santa_2000}.

Our version of the dataset contains sixty single-channel 22,050 Hz .wav files. Each audio file is accompanied by a text-based transcription, in the .cha format, with IU-level timestamps precise to 0.1 seconds. 

Prior to finetuning, IU-level timestamps are extracted from the .cha files which accompany each transcript. Of the 60 transcripts, valid segments of the first five ($\sim$2 hours) are relegated to the testing set, with the remainder allocated to the training/validation sets. Segments are considered valid if they contain no overlap and are connected. Additionally, given the 30-second fixed input length, otherwise valid segments may be split into multiple parts with each containing up to ten consecutive units. Miscellaneous tokens representing other speech artifacts, namely breaths (inhales/exhales) and laughter are removed prior to use. Filled pauses and disfluencies (“um”,”uh”, etc.), however, are preserved.

The extracted timestamps are matched to the input audio and resampled from 22,050 Hz to 16,000 Hz. Log-mel spectrograms of each segment are generated with 80 channels, 25 ms windows, and 10 ms strides. Input matrices are subsequently rescaled to $([0,1])$ and padded to 30 seconds \cite{radford_robust_2022}. 

Manual examination of the preexisting token dictionary is performed to identify tokens which are not desired in a final transcription and occur infrequently so as to minimally disrupt the output. For IU boundaries, we choose the token representing five contiguous exclamation marks.

\subsection{Models}\label{AA}
The transformer is a neural network architecture introduced in 2017 \cite{vaswani_attention_2017}. Unlike recurrent neural networks (RNNs) or convolutional neural networks (CNNs), it applies no structure to the relationship between inputs, both temporally and spatially. Instead, it encodes positional information into the inputs themselves, allowing for computational advantages through parallelization and performance improvements through self- and cross-attention \cite{tay_efficient_2022}. Transformers have achieved state-of-the-art results in a variety of domains, from biology to chemistry, but have found the most success in natural language processing (NLP) tasks \cite{kalyan_ammus_2021}. Whisper \cite{radford_robust_2022}, one such transformer model, manages to achieve competitive results in a variety of speech-processing use cases, including speech-to-text (STT) synthesis.

Our Prosodic Speech Segmentation with Transformers (PSST) model is finetuned from the largest English specific version of Whisper, with 764 million-parameters and a size of 3.06 GB. The two hyperparameter departures from Whisper’s initial training cycle were batch size (256 to 32) and gradient steps (1 to 2) for an effective batch size of 64. These changes solely reflect computational constraints. Two convolutional layers and a Gaussian Error Linear Unit (GELU) activation convert a log mel-spectrogram of the SBCSAE input audio into a linear vector, which is combined with a sinusoidal positional encoding vector. The array is passed through a series of encoder and decoder blocks which are each composed of attention and multi-layer perceptron (MLP) components. Finetuning is conducted in a supervised fashion, using manually generated transcriptions as the ground truth.

We also instantiate a second version of the model (PSST-acoustic), which is trained on a syntax-masked version of The SBCSAE (but otherwise identical). All text tokens are replaced with a common token with boundaries preserved as separators. This is to determine if the PSST model is relying on syntax.

To test whether the PSST-generated boundaries are simply the result of lexical/syntactic probabilities, we train a text-only version of the model. Instead of converting audio to tokens, we task it with placing boundaries between Whisper-generated tokens. Our lexical segmentation model is initiated with the 1.2 billion parameter (5.36 GB) distribution of GPT-NEO \cite{black_sid_gpt-neo_2021} using the tokenizer from GPT-2 \cite{radford_robust_2022}. Training is performed on a text-based version of the pre-processed SCBSAE, with splits identical to those used in PSST and PSST-Acoustic. The basic architecture of PSST, based on \cite{radford_robust_2022}, is shown in figure 1.

\begin{figure}[htbp]
\centerline{\includegraphics[width=90mm]{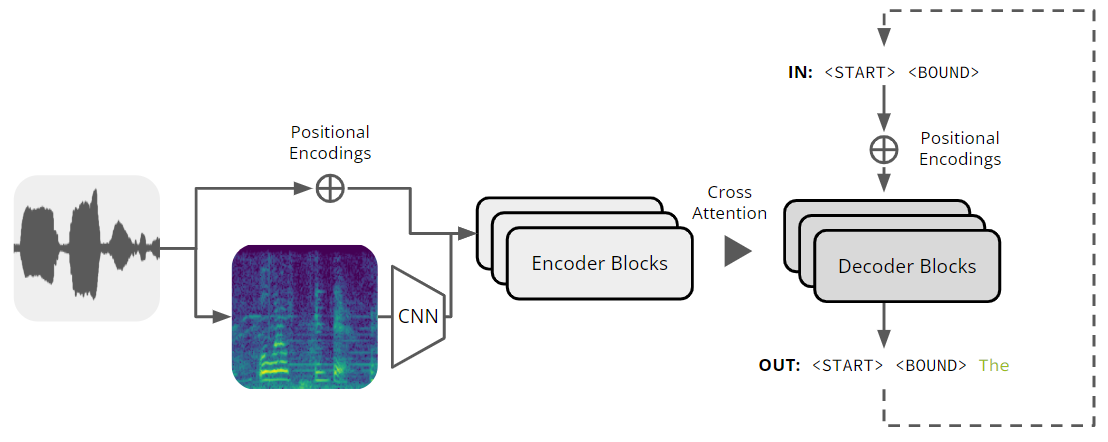}}
\caption{PSST Architecture}
\label{fig}
\end{figure}

\subsection{Evaluation}\label{AA}
Although most ASR systems evaluate model performance based on word error rate (WER) \cite{woodward_information_1982}, as our task involves boundary predictions, our model is evaluated on segmentation metrics. Unlike the STT methods on which we base our model architectures, significant ambiguity underlies the ‘ground truth’ of many prosodic tasks \cite{moore-etal-2016-automated}. Inter-labeler agreement for intonational phrase boundaries, for example, is 93.4\% \cite{pitrelli_evaluation_1994}.

Meaningful segmentation, where proposed boundaries are deemed accurate if their word-level separation agrees with the out of sample expert transcript, are used instead. False positives (over-segmentation errors) and false negatives (under-segmentation errors) tend to occur less often than true negatives, which are found at the remaining non-boundary word to word partitions.  We therefore used accuracy merely as a point of comparison between works. F1 score, computed as the harmonic mean of precision and recall, is preferred as a standard measure of meaningful segmentation performance.
Lexical discrepancies between generated and expertly transcribed portions of audio are resolved through a transformer-based forced alignment technique \cite{zhu_phone--audio_2022}. Cascading errors from both ASR and forced alignment inaccuracies, especially those induced by personal identifier filtering, are partially attenuated by applying a 20 ms alignment window.

\subsection{Training}\label{AA}
The train split is loaded in a streaming fashion for memory purposes. All training occurs on a single NVIDIA V100 Tensor Core GPU with 32 GB of VRAM. 

PSST and PSST-acoustic are trained for 400 steps (2 full passes of the training data). The first 50 steps have a depressed learning rate to avoid early overfitting, with the hyperparameter increasing linearly until it plateaus at $10^{-5}$. This stage requires approximately 2 hours and 20 minutes. 

The lexical model is finetuned on the same hardware as the previous models, with a batch size of eight, 100 warm-up steps, and a 10\% weight decay. Finetuning occurs for two epochs, requiring just under 30 minutes.

\subsection{Inference}\label{AA}
PSST and PSST-acoustic implementation may be performed on CPU, but is significantly accelerated with even consumer grade GPUs. On average, inferences with an NVIDIA T4 GPU take only four seconds per input (up to 30 seconds), with our CPU usually requiring over a minute. The downfolding, resampling, and feature generation steps require comparatively less processing power. Inferences on PSST-lexical require only 1.2 seconds per chunk (1-10 IUs) with GPU acceleration.

\subsection{Signal Reduction Experiments}\label{AA}
Using the PSST and PSST-acoustic models, we apply a series of low-pass and high-pass filters to the audio in the test set. The chosen frequencies of 200 Hz, 400 Hz, 800 Hz, 1.6 kHz, and 3.2 kHz roughly bound the F0-F3 ranges, as noted in \cite{sinha_speech_2010,kent_static_2018}. Evaluation metrics are computed for low-pass and high-pass Butterworth filters with cutoffs at each frequency \cite{butterworth_theory_1930}. As also found in \cite{ferragne_towards_2019}, we observe that upper frequency ranges (beyond 3 kHz) have diminishing effects on speech perceptibility.


\section{Results}
\subsection{Performance}\label{AA}
Our method achieves state-of-the-art performance in both F1 score and overall accuracy, even over some methods with an accompanying human-labeled orthography. However, variation in segment definitions, input features (syntactic and/or acoustic), corpus content (number of speakers, scripted or unscripted, etc.), and model type make comparisons difficult. Table 1 summarizes the performance of previous English-specific segmentation methods.

\begin{table}[th]
\caption{Segmentation Performance}
\centering
\begin{tabular}{lrr}
\toprule
\emph{Method} &    \emph{F1} &  \emph{Accuracy} \\
\midrule
\textbf{PSST}       &  \textbf{0.87} &      \textbf{0.96} \\
\cite{rosenberg_automatic_2009} &  0.81 &      0.93 \\
\cite{rosenberg_classification_2010} &  0.77 &      0.89 \\
Whisper \cite{radford_robust_2022} + Lexical &  0.77 &      0.93 \\
PSST-Acoustic &  0.71 &      0.87 \\
\cite{hirschberg_acoustic_1998} &  0.70 &      0.83 \\
\cite{biron_automatic_2021}      &  0.66 &      0.86 \\
\cite{klejch_punctuated_2016}     &  0.63 &      0.87 \\
Whisper \cite{radford_robust_2022}  &  0.48 &   0.85 \\
\bottomrule
\end{tabular}
\end{table}

\smallskip{}

We also compare the distributions of IU length, finding out-of-sample similarity between PSST-generated and manually transcribed IU densities. The distributions are shown in fig. 2.

\begin{figure}[htbp]
\centerline{\includegraphics[width=90mm]{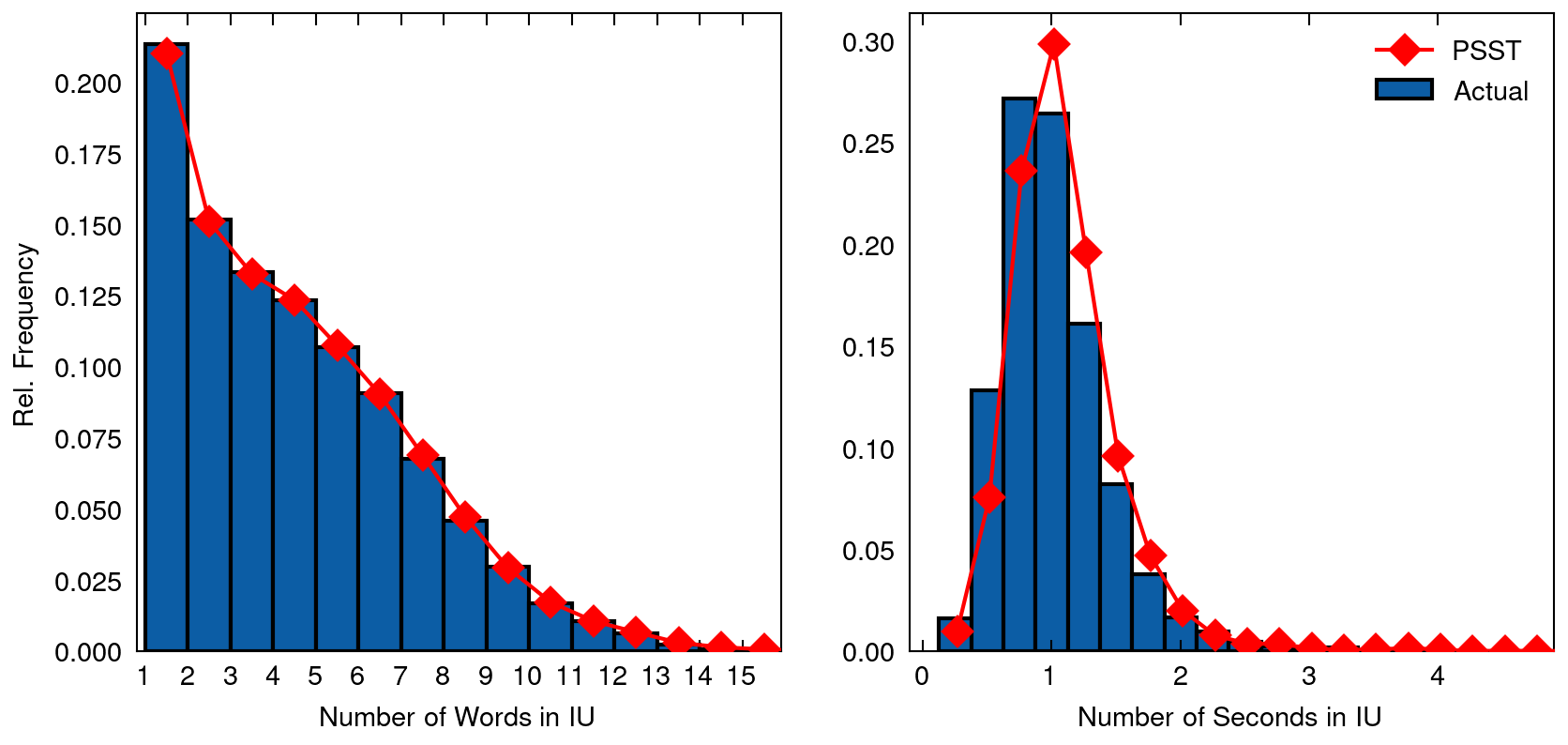}}
\caption{IU Length Distributions}
\label{fig}
\end{figure}

\subsection{Performance on IViE Corpus}\label{AA}
We test PSST on the Intonational Variation in English (IViE) corpus \cite{grabe_modelling_2001}. Unlike The SBCSAE, IViE focuses on urban dialects of English spoken in the British Isles, and is transcribed with a distinct intonational phrase methodology. The IViE labeling system is adapted from the ToBI framework \cite{silverman1992tobi, beckman1997guidelines}. Despite this, we find robust yet degraded performance in this out-of-distribution environment. Table 2 summarizes performance on the IViE corpus.

\begin{table}[th]
  \caption{IViE Corpus}
  \centering
    \smallskip{}
    \begin{tabular}{lrrrr}
    \toprule
    \emph{Method} &    \emph{F1} &  \emph{Accuracy}\\
    \midrule
    \textbf{PSST} &  \textbf{0.73} & \textbf{0.93}\\
    Whisper \cite{radford_robust_2022} + Lexical &  0.56 & 0.89\\
    Whisper \cite{radford_robust_2022} & 0.35 & 0.87 \\
    PSST-Acoustic &  0.00 & 0.82\\
    \bottomrule
    \end{tabular}
\end{table}

\subsection{Failure Cases}\label{AA}
Failure cases fit into two broad categories: ASR-induced inaccuracies and prosodic inaccuracies. The STT portion of our model generates some tokens which are not included in the expert transcription, and fails to generate others. We find this to be especially problematic with barely-audible filled pauses. Given that such tokens are often associated with boundaries, these cases detract from PSST’s overall performance. Upon listening to the audio in question, we note ambiguity in the existence of these vocalizations and if their existence would warrant additional IUs.

Another form of ASR-induced inaccuracies are those which output longer or shorter tokens, which cannot be aligned within the 20 ms window. Lengthening the alignment window would reduce these cases at the cost of potentially marking inaccurate boundaries as correct.

In cases of correct or near-correct lexical transcriptions, IU segmentation errors are more clearly attributable to prosodic factors. We observe these errors to accompany more subjective examples.

\subsection{Filters}\label{AA}
We find a slight ($\sim$0.1\%) improvement in segmentation performance after applying a 3.2 kHz high-pass filter, while performance reductions accompany all other frequencies and filters. More extreme filters are associated with larger reductions in overall performance, as shown in fig. 3. 

\begin{figure}[htbp]
\centerline{\includegraphics[width=65mm]{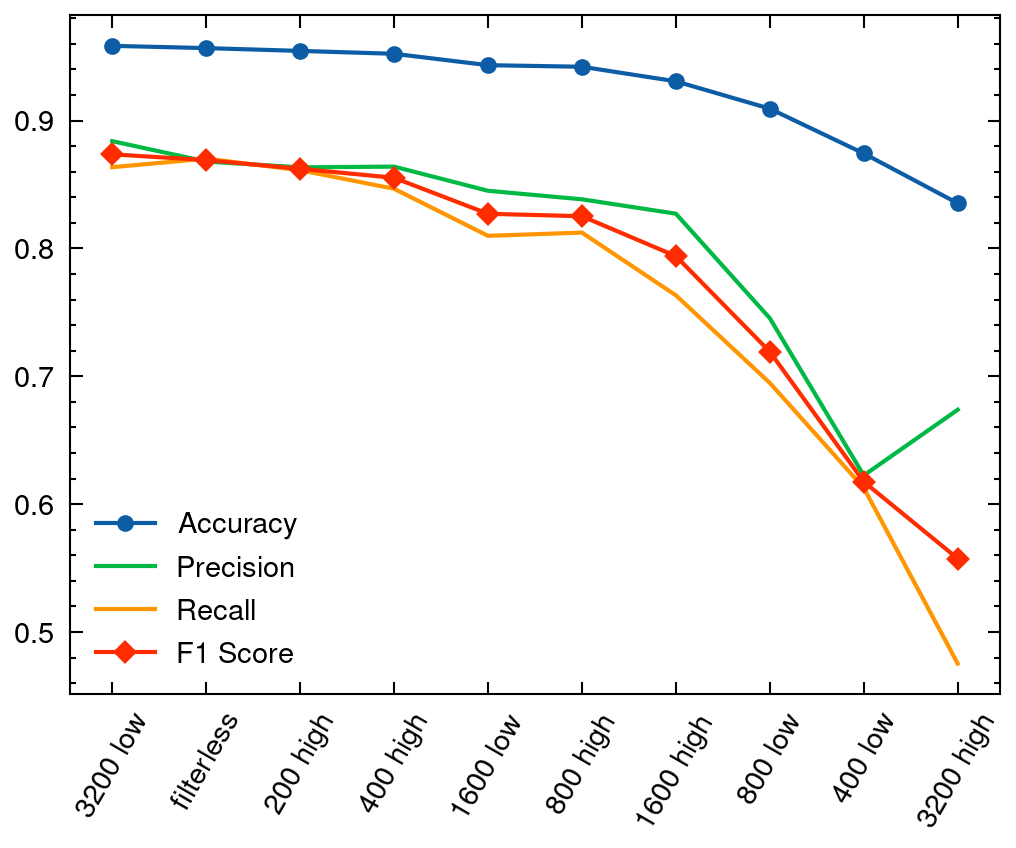}}
\caption{Filter Frequency and Type}
\label{fig}
\end{figure}

Although we evaluate segmentation metrics, STT accuracy presents endogenous effects. Segments with complete orthographic agreement also  yield strong IU boundary agreement. It is therefore difficult to determine whether more central harmonic frequencies capture significant prosodic information or simply orthographic clues.

\section{Discussion \& Conclusion}
The study set out to accomplish three research objectives. In relation to Objective 1, we successfully repurposed Whisper \cite{radford_robust_2022} to segment spontaneous speech into IUs. We achieved an F1 score of 0.87 on previously unseen examples, higher than any legacy method. Whisper was originally trained on the simple objective of discerning words from audio, yet the fact that we were able to repurpose it successfully using few-shot learning holds significant promise for other NLP studies that rely on smaller datasets.

In relation to Objective 2, we investigated the performance of two `sibling` models, finetuned on a lexical version and an orthographically-confounded (PSST-acoustic) version of the SBCSAE. The full PSST model performed substantially higher than both the lexical model and the acoustic model, which achieve F1 scores of 0.77 and 0.71 respectively. These results confirm that both prosody and syntax have a role to play in the determination of boundaries.

In relation to Objective 3, the results indicated large deltas in performance on in-distribution and out-of-distribution datasets. When applied to the out-of-distribution IViE dataset, the PSST model was successful in predicting intonation boundaries with an F1 score of 0.731. Intonation variation between accents, such as those of the British Isles can be wide, as reported by Grabe et al., 2001. It is therefore expected that there are likely more significant differences between American and British English dialects \cite{dichristofano_performance_2022}. It is very promising that the model, which was trained on the SBCSAE dataset, was able to achieve this level of accuracy in predicting boundaries in other dialects of English. However, we suspect optimal performance will involve a finetuning set which includes multiple varieties of English, including those with distinct L1 influence. Similarly, performance discrepancies were found in distinct harmonic filtering environments, with notable declines in performance following sub-800 Hz and super-1600 Hz masks. The 200-1600 Hz range, roughly corresponding to F1 and F2 in English, contained the most useful information for the prosodic segmentation task \cite{catford_practical_1988}. This result was unexpected, given the prominence of F0 in intonation.

Taken together, on the basis of this research we postulate that text prediction and prosodic boundary identification are not independent challenges, but merely components of a unified speech processing objective. Simply re-tokenizing prosodic features in a manner that transformer-based models can process unlocks a seemingly latent ability to identify IUs. Overall, our results suggest that such STT models implicitly consider prosody, given their success in a few-shot context. Furthermore, the robustness of segmentation performance when exposed to moderate frequency-based signal tampering, or even complete F0 masking, strengthens the case for prosody-syntax interplay at the “heart” of high-performance ASR models. 

Future work may consider our ASR retokenization process to detect other speech phenomena, such as prosodic accents, vocal quality changes, or even environmental contexts. 

\section{Acknowledgments}

We thank Dr. John DuBois for provisioning a copy of The SBCSAE and Dr. Tirza Biron for assistance in determining robust performance metrics. This work was funded by an URCA grant from the University of California, Santa Barbara. 

\bibliography{references}{}
\bibliographystyle{plain}
\end{document}